\documentclass[conference]{IEEEtran}
\IEEEoverridecommandlockouts
\usepackage{cite}
\usepackage{amsmath,amssymb,amsfonts}
\usepackage{algorithmic}
\usepackage{graphicx}
\usepackage{textcomp}
\usepackage{amsmath,amssymb}
\DeclareMathOperator{\E}{\mathbb{E}}
\usepackage{svg}
\usepackage{xcolor}
\def\BibTeX{{\rm B\kern-.05em{\sc i\kern-.025em b}\kern-.08em
    T\kern-.1667em\lower.7ex\hbox{E}\kern-.125emX}}
\begin{document}

\title{A Survey on Reinforcement Learning for Combinatorial Optimization
}

\author{\IEEEauthorblockN{Yunhao Yang}
\IEEEauthorblockA{\textit{Department of Computer Science} \\
\textit{Univerisy of Texas at Austin}\\
Austin, USA \\
yunhaoyang234@utexas.edu}
\and
\IEEEauthorblockN{Andrew Whinston}
\IEEEauthorblockA{\textit{McCombs School of Business} \\
\textit{University of Texas at Austin}\\
Austin, USA \\
abw68@utexas.edu}

}

\maketitle

\begin{abstract}
This paper gives a detailed review of reinforcement learning (RL) in combinatorial optimization, introduces the history of combinatorial optimization starting in the 1950s, and compares it with the RL algorithms of recent years. This paper explicitly looks at a famous combinatorial problem---traveling salesperson problem (TSP). It compares the approach of modern RL algorithms for the TSP with an approach published in the 1970s. By comparing the similarities and variances between these methodologies, the paper demonstrates how RL algorithms are optimized due to the evolution of machine learning techniques and computing power. 
The paper then briefly introduces the deep learning approach to the TSP named deep RL, which is an extension of the traditional mathematical framework. In deep RL, attention and feature encoding mechanisms are introduced to generate near-optimal solutions. The survey shows that integrating the deep learning mechanism, such as attention with RL, can effectively approximate the TSP. The paper also argues that deep learning could be a generic approach that can be integrated with any traditional RL algorithm to enhance the outcomes of the TSP.
\end{abstract}


\section{Introduction}

We focus on the potential solutions for combinatorial optimization, also known as discrete optimization, as opposed to continuous optimization. Discrete optimization searches for an optimal solution in a finite or countably infinite set of potential solutions. Optimality refers to a global maximum or minimum of a certain criterion function.
This paper discusses combinatorial optimization applied to the quadratic assignment problem (QAP) and a special case of it, which is the well-known traveling salesperson problem (TSP).  

TSP is an NP-hard problem, and finding the optimal solution requires high computational complexity. Hence developing a low-complexity algorithm to estimate the solution is essential. Therefore, many researchers focus on the approximation algorithms of the TSP, which could largely reduce the computational complexity. 

Recently, researchers have been focusing on applying RL algorithms to approximate the solution to the TSP.
Given the evolution of computing power, RL algorithms are capable of solving problems in more aspects. Unlike supervised learning, RL does not need labeled data to adjust the network based on the loss function. Instead, it finds a balance between exploration and exploitation \cite{ka}. 
RL was not developed to find the optimal solution but rather to approximate it. Therefore, RL is a new approach worth exploring compared to approximating TSP arithmetically. 

We survey RL-related approximation algorithms following a historical timeline. We discuss the performance differences between the algorithm introduced in the 1960s and modern RL algorithms and how computing power limitations can affect their performance.
With deep learning development, neural networks are integrated with RL. Neural networks can enhance the learning ability of RL. In terms of the TSP, modern techniques such as deep RL or the attention mechanism \cite{attention} can also approximate this NP-hard problem. We explore and compare those deep learning techniques with the previous methods.

\section{Preliminary}
\paragraph{Reinforcement Learning}
(RL) is a machine learning technique that develops approximate methods for solving dynamic optimization problems. 
To distinguish RL from other machine learning techniques, an RL agent is not given the optimal action. Hence the agent needs to try different actions and finds the optimal one \cite{rlbook}. 
Therefore, the RL agent's main concern is to decide which action is optimal. An optimal action leads to the highest reward obtained from the environment, which is typically represented by a Markov decision process \cite{va}. Because of the nature of RL, it is a relatively efficient learning technique. 

RL algorithms typically consist of an agent that takes actions. The actions are determined based on environmental observations and policy. Each new action updates the environment and generates new observations. This loop continues until it reaches a terminating state (e.g., game over) or the maximum number of steps allowed. The rewards are given based on the outcomes of the actions. The learning policy is optimized to maximize the cumulative reward. The RL policy is the component that we want to learn. The policy is a function-like component that takes into account the environmental observations and outputs the expected actions. 

The Q-learning algorithm is a traditional RL algorithm that consists of a Q-value table, and the policy takes action with the highest Q-value. As the RL model becomes applicable to more complicated problems, the policy becomes more complex as well. 

Actor-critic is another popular RL algorithm that has two components: An actor and a critic \cite{ac}. An actor is the policy function that evaluates the current state and determines the action. 
A critic is the value function that evaluates the actions made by the actor. Typically, an RL agent takes action to reach a state, and the critic estimates the expected reward that the agent can get from this state.

In recent years, the concept of deep RL was introduced and applied in the field of machine learning. Deep RL is a combination of RL and deep learning \cite{flt}. Deep RL utilizes a deep neural network structure to manipulate high-dimensional data. It typically generates output based on probabilistic outcomes. Deep RL algorithms have been used in many fields \cite{YANG2022537, Bajaj2021ReinforcementLO}, including generating approximations to the TSP and other combinatorial optimization problems. 

\paragraph{Quadratic Assignment Problem}
(QAP) is a problem of locating indivisible economic activities \cite{an}. The QAP consists of two sets of interrelated objects and the solution to the problem is the optimal assignment among the objects. From an economic perspective, the goal of the QAP is to allocate a collection of facilities to a list of locations in a manner that effectively minimizes the overall cost associated with the assignments \cite{ce}. 

The QAP is known to be an NP-hard problem \cite{ce}. There is no polynomial-time solution to the problem; however, many approximation algorithms for this problem reduce its computational complexity. Recently, the idea of machine learning has been broadly used in computation. Many researchers have currently been trying to apply RL and neural networks to solve the QAP,  an important way that we discuss later in the paper. 

\paragraph{Traveling Salesperson Problem}
(TSP) is a classic combinatorial optimization problem. The problem provides a set of cities and their positions and asks for the shortest path to visit all the cities exactly once.
Formally, given a set of cities and the distances between each pair of cities, the goal is to determine the shortest path for visiting all cities exactly once. 
It is an NP-hard problem, meaning that any other NP problem can be reduced to it using a polynomial-time transformation.

\section{Historical Timeline}
Although RL is a relatively new field in machine learning, researchers have introduced the idea of using RL to solve the TSP. The idea of RL can be traced back to mid 20th century, providing theoretical support for modern RL. For example, a prototype of RL in combinatorial optimization was introduced in 1970. 

\subsection{Dynamic Programming (1950s)}
Dynamic programming (DP) refers to simplifying a decision by breaking it down into a sequence of decision steps. This is done by defining a value table. Each step determines the value by looking at the previously stored values. The DP algorithm observes previous values as the state and then determines the action---or current value---based on the state. The current value is stored in the value table and will be used as a part of the state for the next steps. In machine learning, RL algorithms borrow DP's state-action idea.

DP is often applied to solve NP problems. A common application is in generating a Fibonacci sequence. To find the $n$th Fibonacci number, we can create a table to store the previous two Fibonacci numbers, add them together to obtain the current number $f_n = f_{n-1} + f_{n-2}$, and replace $f_{n-2}$ with $f_n$.

DP is also capable of solving NP-hard problems. A common example is the knapsack problem. There are also DP algorithms for solving the TSP. As an example, one DP algorithm divides the TSP into multiple sub-problems. The agent computes the optimal distance for each sub-problem and returns the optimal distance to the main branch. 

However, even though DP can reduce the computational complexity in most cases, DP for the TSP is still NP-hard. The algorithm computes sub-problems recursively. Each sub-problem is of $O(n)$ complexity, and there are a total of $n2^n$ sub-problems. Therefore, the overall computational complexity of this DP algorithm is $O(n^2 \cdot 2^n)$, which is in exponential time. This indicates that the DP algorithm can also not reduce the complexity to polynomial time. Therefore, we want to explore RL algorithms that are inspired by DP but able to approximate the results in polynomial computational time.

In general, if sub-problems can be nested recursively inside larger problems (as in recursions in computer science), DP methods are applicable. Then, there is a relation between the larger problem's value and the sub-problems \cite{Cormen}, and such a relationship can be computed by the Bellman equation.

\subsection{Bellman Equation (1957)}
The Bellman equation, also known as the DP equation, was introduced by Richard Bellman \cite{bell} and has been used in DP. It divides a dynamic optimization problem into a sequence of simpler subproblems \cite{kir}. The Bellman equation can address most discrete-time problems related to optimal control theory. By contrast, the Hamilton--Jacobi equation addresses the continuous case \cite{hje}. The Bellman equation was inspired by the Hamilton--Jacobi equation and was introduced for discrete problems. 

The Bellman equation is commonly used as the starting point of the RL approach. We can use two types of value functions to learn the optimal policy $\pi$ in RL \cite{graves}: the state value function $V(s)$ or the action value function $Q(s, a)$. 
The state value function returns the value of a state $s$ according to policy $\pi$ (a function or method to generate outputs): 
\begin{equation}     
    V^{\pi}(s) = \E_{\pi}[R_{t}|s_{t} = s] 
\end{equation} 

Define $\wp_{s s'}^{a} = Pr[s_{t+1} = s' | s_{t} = s, a_{t} = a]$ and $\Re_{s s'}^{a} = \E[r_{t+1}| s_{t}=s, s_{t+1}=s', a_{t} = a]$, where $\wp$ is the transition probability and $\Re$ is the expected or average reward when starting in state $s$, taking action $a$, and moving toward state $s'$. Then, to derive the Bellman equation, we can write the state value function as 
\begin{equation}
    \begin{aligned}
        V^{\pi}(s) &= \E_{\pi}[\sum_{k=0}^{\infty} \gamma^{k}r_{t+k+1}|s_{t}=s] \\ &= \E_{\pi}[\gamma_{t+1} + \gamma\sum_{k=0}^{\infty} \gamma^{k}r_{t+k+2}|s_{t}=s]
    \end{aligned}
\end{equation}
This equation describes the expected return value when starting in state $s$ and following policy $\pi$. Then, we insert $\wp$ and $\Re$, which are defined above, and use the fact that 
\begin{center}
    $\E_{\pi}[r_{t+1}|s_t = s] = \sum_a \pi(s, a) \sum_{s'}\wp_{s s'}^{a}\Re_{s s'}^{a}$
\end{center}

\begin{equation}
    \begin{aligned}
        \E_{\pi}[\gamma\sum_{k=0}^{\infty} & \gamma^{k}r_{t+k+2}|s_{t}=s] = \\ & \sum_a \pi(s, a)\sum_{s'}\wp_{s s'}^{a}\gamma\E_{\pi}[\sum_{k=0}^{\infty} \gamma^{k}r_{t+k+2}|s_{t+1}=s']
    \end{aligned}
\end{equation}

Eventually, the state value function $V(s)$ can be rewritten as 
\begin{equation}
    V^{\pi}(s) = \sum_a \pi(s, a)\sum_{s'}\wp_{s s'}^{a}(\Re_{s s'}^{a} + \gamma V^{\pi}(s'))
\end{equation}

The action value function can also be derived by using the Bellman equation as follows: 

\begin{equation*}\label{eq:pareto mle2}
  \begin{aligned}
  Q^{\pi}(s, a) &= \E_{\pi}[\sum_{k=0}^{\infty} \gamma^{k}r_{t+k+1}|s_{t}=s, a_t=a] 
    \\&= \sum_{s'}\wp_{s s'}^{a}(\Re_{s s'}^{a} + \gamma \sum_{a'} \pi (s', a') Q^{\pi} (s', a'))
 \end{aligned}
\end{equation*}

The action value function can be rewritten as:
\begin{equation}
    Q^{\pi}(s, a) = E_{\pi}[R_t|s_t=s, a_t=a]
\end{equation}

Therefore, the Bellman equation allows one to express a specific state's values as those of other states. This simplifies the calculation of values between states, opening up many possibilities for iterative approaches to calculating the value for each state \cite{graves}. The Bellman equation thus played an important role in the inception of RL. However, the Bellman equation cannot solve large-scale NP-hard problems such as the TSP. RL provides a way to approximate the Bellman equation and solve the TSP. 

\subsection{Quadratic Assignment Algorithm (1970)- A Prototype of RL} The quadratic assignment algorithm for solving the TSP was introduced in 1970 \cite{wh}. This algorithm uses the Bellman equation and statistical properties of the criterion function. It can be considered a prototype of RL that calculates the mean and variance to serve as a value function. The quadratic assignment algorithm achieved one of the shortest distances for the TSP among existing algorithms in that period. 

\subsubsection{Computational Scheme}
The TSP is considered an optimal permutation problem. The algorithm attempts to discover the optimal mapping between S and R sets, where S consists of variables $(x_1, ..., x_n)$ and R consists of integers from $(1, ..., n)$. \newline The process is as follows: 

\begin{enumerate}
    \item Initialize by setting k = 1 and determine $S_1$ and $R_1$ according to a specific rule (eg. feasibility).
    \item If $R_k$ is empty, go to step 8.
    \item Take elements $i^* \in S_k$ and $j^* \in R_k$ arbitrarily. Let $i_k = i^*$, $j_k = j^*$, and $R_k = R_k\backslash \{j^*\}$.
    \item If k = n, go to step 12.
    \item Increment k by 1, $k = k + 1$.
    \item Determine $S_k$ and $R_k$ according to some specified rule.
    \item Repeat from step 2.
    \item If k = 1, stop
    \item Decrement k by 1, $k = k-1$.
    \item If $R_k$ is empty, go back to step 8.
    \item Take an arbitrary element $j^* \in R_k$, let $j_k = j^*$ to go with the current $i_k$ and set $R_k = R_k\backslash \{j^*\}$. Then go back to step 4
    \item Record the current mapping, and go to step 10.
\end{enumerate}

This algorithm involves the value of the associated mean completion as a criterion for selecting a path: 
\begin{center}
    $\E[\phi(p)$, $p \in C(i_1, ...i_k)]$
\end{center}

For the $k^{th}$ selection, the algorithm will pick the path to the city that has the smallest mean among all unvisited cities. The mean can be calculated as

\begin{center}
    $M = \frac{1}{n} (\sum_{i \in N a_i}) (\sum_{j \in N b_j})$
\end{center}

And the variance can be written as:
\begin{center}
    $\frac{1}{n!} \sum_p [(\sum_j a_j b_{p(j)}) (\sum_i a_i b_{p(i)})]$
    \newline
    $= \frac{(n-2)!}{n!} [(\sum_i a_i)^2 - \sum_t a_t^2][(\sum_k b_k)^2 - \sum_t b_t^2] + \frac{(n-1)!}{n!}(\sum_i a_i^2)(\sum_j b_j^2)$
\end{center}

Repeated examinations with the selection procedure indicate that, while an optimal solution does not generally result from the first complete assignment, a very good solution is achieved. The algorithm then turns to the explicit computation of the mean and variance values of the completion of a $k$-partial map. 

\subsubsection{Implicit Enumeration}

Define $\Psi(S)$ as a set that consists of all the subsets of the set S. Let $L_{\phi}: \Psi (S) \rightarrow R$ and $L_{\phi} (E) \le \phi (i), \forall i\in E$ where $E \subset S$. The functions give a lower bound for $\phi (i)$ over the subset E. Then, let $A$ denote the value of the current best-completed assignment, where $A = \infty$ if there is no complete assignment yet. If $L_{\phi} (C(i_1, ..., i_k)) \ge A$, there is no $k$-partial mapping from $i$ to $j$ that consists of a better solution than the current mapping. The completion class $C(i_1,...,i_k)$ is {\it implicitly enumerated}. 

To embed implicit enumeration into the algorithm, replace step 11 with the following: 

\begin{center}
{\it 11. Take an arbitrary element $j^* \in R_k$, let $j_k = j^*$ to go with the current $i_k$ and set $R_k = R_k\backslash \{j^*\}$. Compute $L_{\phi} (C(i_1, ..., i_k))$. If $L_{\phi} (C(i_1, ..., i_k)) \ge A$ go to step 9, otherwise go back to step 4.}
\end{center}

Define $G_{\phi}: P(S) \rightarrow R$ as the $\alpha$-probabilistic lower bound set function for any given $\phi$ and $\alpha$. In terms of the overall algorithm, the $\alpha$-probabilistic lower bound functions $G_{\phi}$ are employed as the lower bound functions $L_{\phi}$ in step 11 presented above. This means that if 
\begin{center}
    $G_{\phi} (C(i_1,...,i_k)) \ge A$,
\end{center}
then completion class $C(i_1,...,i_k)$ is implicitly enumerated at the $\alpha$ confidence level. The computational experience presented in the paper indicates that the $\alpha$-probabilistic lower bound function provides much greater cutting power than the lower bound functions without substantial risk of overlooking the true minimum. It has been proved to achieve complete confidence level enumeration with large practical problems. 

\subsubsection{Evaluation}
The algorithm is illustrated for three classical TSPs from Karg and Thompson \cite{karg}: the 33-, 42-, and 57-city problems, which are considered large-scale problems. The outcomes of the quadratic assignment algorithm and the optimal solution of the problems are presented in the table below. The optimal solution is retrieved from \cite{karg}. The results show that the approximations are within 15\% of the optimal solution; moreover, the approximations will approach the optimal solution as the problem scales up. 


\subsubsection{Analysis}
The quadratic assignment algorithm exhibits some of the characteristics of RL. It utilizes the Bellman equation to approximate the optimal solution of the partial TSP. The Bellman equation is the basic block for solving RL and is omnipresent in modern RL \cite{tan}. In the quadratic assignment algorithm, the Bellman equation is employed to calculate the optimal solution of the prevailing $k$-partial TSP. The calculation is built upon the recorded optimal solution of the previous ($k - 1$)-partial TSP. Then, the determination of $k$-partial mapping will serve as the environment for the subsequent ($k+1$)-partial mapping. The mean and variance serve as the value function in the modern RL architecture. The key to solving the TSP is to minimize the mean distance of each $k$-partial mapping. Therefore, a large mean value will diminish the probability of a specific choice at the current step. 

Unmistakably, many distinctions exist between Grave and Whinston’s approach and modern RL due to the inadequate computing power in the 1960s. There is no adjustment of the network in the training procedure in Graves and Whinston's architecture. The learning component in the model is not significant. 

\subsection{Ant-Q (1995)- A Classical RL Approach}
The ant-Q algorithm was introduced by Luca M. Gambardella and Marco Dorigo in 1995, and it presents many similarities with Q-learning \cite{wat}. The ant-Q algorithm was inspired by the ant system \cite{as}, a distributed algorithm for combinatorial optimization. It was a notable algorithm that applied RL to the TSP. The ant-Q algorithm has obtained competitive results for both symmetric and asymmetric TSPs \cite{antq}. 

\paragraph{Ant System (AS)} is the first ant colony optimization algorithm \cite{antq}, which reduces computational problems to path search problems. The ant system uses an artificial ant---a computational agent---to find good solutions to graph-related optimization problems. The optimization problem to which we can apply ant colony optimization is equivalent to the TSP, which finds the shortest path on a weighted graph. In the ant colony optimization algorithm, each artificial ant selects a path to arbitrarily construct a solution for every iteration. Then, the solutions generated by the ants are compared and evaluated. The network is then adjusted based on the evaluations. 

\paragraph{Q-Learning} is an RL algorithm that learns and records policies and takes action to maximize the expectation. The letter $Q$ stands for the quality of the actions that are taken. Q-Learning is model-free, so it does not require environmental networks to be built.  

The algorithm is centralized on the quality of state--action combinations: 
\begin{center}
    $Q: S\times A \rightarrow \mathbb{R}$
\end{center}

In the beginning, $Q$ is initialized randomly. At each time step $t$,  action $a_t$ is associated with a transformation cost/reward to the next stage $s_{t+1}$, where the cost/reward is denoted as $r_t$. The quality $Q$ is updated by utilizing the Bellman equation as the following: 

\begin{center}
    $Q_{new}(s_t, a_t) = Q_t(s_t, a_t) + \alpha (r_t + \lambda \cdot max \{ Q(s_{t+1}, a)\} - Q(s_t, a_t))$
\end{center}
where $\alpha$ is the learning rate and $\lambda$ is the discount factor.

\paragraph{Ant-Q Algorithm} integrates Q-learning with the ant system. Each iteration involves the following four steps: 
\begin{itemize}     
    \item Initialize the AQ values and place each agent $k$ on a city $r_{k1}$ according to some policy. Initialize a set of cities $J_k(r_{k1})$ to be visited.     
    \item Each agent makes a move and updates $AQ(r,s)$ if the move discounts the next state evaluation. The agents continue moving and updating the AQ values until they are back in the starting city.     
    \item The length $L_k$ of the tour of agent $k$ is computed, and $L_k$ is used to calculate the delayed reinforcements $\Delta AQ(r, s)$. Then, the AQ-values are updated based on $\Delta AQ(r, s)$.     
    \item Check whether the predefined termination condition is met. Return the approximated shortest path $L_k$.
\end{itemize} 
The AQ values are updated by the following rule:
\begin{center}
    $AQ(r,s) = (1 - \alpha)\cdot AQ(r,s) + \alpha \cdot (\Delta AQ(r,s) + \lambda \cdot Max_{z \in J_k(s)} AQ(s,z))$
\end{center}
where $J_k(s)$ is a function of the previous history of agent $k$. Then $\Delta AQ(r,s)$ is calculated as follows: 
\begin{center}
    $\Delta AQ(r,s) = 
    \begin{cases}
               \frac{W}{L_k} \text{ if } (r,s) \in \text{tour done by agent }k\\
               0 \text{ otherwise}
    \end{cases}$
\end{center}

\subsubsection{Evaluation}
The approximation results of the Ant-Q algorithm are compared with the following approaches. 
\begin{itemize}     
    \item Elastic Net: a regression method with L1 and L2 regularization on the lasso and ridge methods.
    \item Simulated Annealing: a probabilistic technique for approximating the global optimum of a given function. In the TSP \cite{car}.     
    \item Self-organizing Map: a self-supervised neural network.
\end{itemize} 

The comparison uses five 50-city problems. The results show that the ant-Q algorithm achieves the minimum average distance among all the approaches in four of the five problems. This indicates that RL does display good performance in the TSP. 

\subsubsection{Analysis}
The ant-Q algorithm is a typical RL algorithm that can be effectively applied to the TSP. It stores the AQ-values and utilizes them to determine the optimal path. This is an effective model-free RL algorithm. Therefore, the computations are not as complicated as for the algorithms with network models. Compared to the Quadratic Assignment Algorithm, the ant-Q algorithm performs better on medium-size maps (e.g., 50 cities). 

However, the limitation of Q-learning also applies to the ant-Q algorithm. A model-free algorithm is not compatible with a large number of environmental factors. The AQ values are the only determining factors of the paths, so the algorithm is not able to take many states/actions into consideration. Therefore, the algorithm could eliminate a path that is costly in the short run but efficient in the long run. 

The limitation of Ant-Q is no longer a problem when the concept of deep RL is introduced. Deep RL is capable of scaling to previously unsolvable problems\cite{aru}. 

\subsection{REINFORCE (2019)- A Deep RL Approach}
Deep RL involves one or more deep neural networks as its policy. Each neural network contains multiple matrices for generating good policy. The parameters of the neural networks are optimized during the training procedure to maximize the reward. Intuitively, the RL policy is learned from the training dataset. The parameters in the policy memorize the experiences. Appropriate losses and gradient descent methods are chosen to optimize the parameters so that the policy will take ``long shots" into consideration. One example of a deep RL algorithm is called REINFORCE. 

The REINFORCE algorithm presents the idea of learning a heuristic for combinatorial optimization problems. 
The model's training process employs a simple benchmark based on a deterministic greedy rollout, which is more efficient compared to a value function \cite{kool}. 
This algorithm significantly improves the performance of the TSP in up to 100 cities. 

\subsubsection{Attention Model}
The attention model takes the graph structure into account through a masking procedure. The algorithm implements an encoder-decoder model with a standard multi-head attention mechanism embedded. The model produces a solution $\pi$ through a stochastic policy $p(\pi | s)$, where $s$ is the problem instance. The policy is factorized and parameterized by $\theta$:
\begin{center}
    $p_\theta(\pi | s) = \prod_{t=1}^n p_{\theta}(\pi_t|s, \pi_{1:t-1})$.
\end{center}

Deep neural networks require feature embedding, a numerical representation of raw data. In this algorithm, the encoder returns  embeddings of all the input nodes. For a TSP as a graph with $n$ nodes, an input node is defined as a node $n_i$ for $i = 1,...,n$, represented by a feature $x_i$ to indicate its connections to other nodes. 

In particular, the encoder consists of $N$ sequential layers, and each layer is composed of a fully-connected layer with multi-head attention. The encoder takes graph nodes as input and returns the embeddings for them. Such embeddings are fed into the decoder, in which the decoder returns a solution $\pi$.

To produce the solution, a context vector that consists of the graph embedding of the first, last, and unvisited cities will be given to the decoder. The decoder will calculate the probability distribution of unvisited cities and output the next city to be visited. We can determine the next city to be visited based on the probability distributions. We can thus construct a path of the TSP. 

\subsubsection{Algorithm}
The attention model obtains a solution (path) $\pi |s$ from a probability distribution $p_\theta(\pi | s)$. To train the model, we define the loss as  
\begin{center}
    $L(\theta | s) = \E_{p_\theta}[L(\pi)]$
\end{center}
The algorithm minimizes loss by gradient descent, using the REINFORCE gradient estimator with baseline $b(s)$\cite{will}:
\begin{center}
    $\Delta L(\theta | s) = \E_{p_\theta}[(L(\pi) - b(s)) \Delta \textit{log } p_\theta (\pi |s)]$.
\end{center}
The model is thus trained to improve itself. Intuitively, the REINFORCE algorithm optimizes the parameters in the attention model based on the length of the path that is computed. A longer path length indicates that this path is far from the optimal path, and the attention model is thus required to adjust itself to generate a better path. 

\subsubsection{Evaluation}
For the TSP, the algorithm is compared with the nearest insertion, random insertion, farthest insertion, as well as nearest neighbor heuristics. 
\begin{itemize}
    \item Nearest Insertion: $i^* = arg min_{i \notin S} (\min_{j \notin S} (d_{ij}))$.
    \item Farthest Insertion: $i^* = arg max_{i \notin S} (min_{j \notin S} (d_{ij}))$.
    \item Random Insertion: inserts a random node.
    \item Nearest neighbor: represents the partial solution as a path with a starting and an ending node. 
\end{itemize}
These heuristics are compared using the 20-, 50-, and 100-city TSP, while the REINFORCE algorithm obtains the best performance among all the selected algorithms within a relatively short period. 

\subsubsection{Analysis}
The REINFORCE algorithm presents an approach to utilizing RL techniques for self-training the attention model or graph attention network. The attention model provides a probabilistic approach to estimating the path based on the given environment. The attention model focuses on solving subproblems and merging the outcomes probabilistically to form the final result. It can thus reduce the overall computational complexity by adjusting the parameters in the attention network to set concentrations (a subgraph with high probability scores). Compared to the Quadratic Assignment Algorithm and the ant-Q algorithm, REINFORCE performs significantly better due to the utilization of deep neural networks.

REINFORCE integrates deep neural networks with RL. A roll-out network is used to deterministically estimate the difficulty of the instance and periodically updates the roll-out network with the parameters of the policy network \cite{riv}. REINFORCE tries to optimize the problem by optimizing the parameters in the attention model. This algorithm achieves high performance in small-scale problems (problems of up to 100 cities) but cannot solve very large problems. The deterministic greedy roll-out reduces the complexity of the problem. However, as the scale of the problem increases, the roll-out will limit the potential to approach the optimal solution. 

More recently, researchers have trained graph convolutional networks using a probabilistic greedy mechanism to predict the quality of a node and embed a Q-learning framework into the network \cite{manc}. This more recent algorithm largely reduces the computational complexity and is capable of million-city problems. 

\section{Conclusion}
Through analyzing three RL approaches to the TSP, we argue that RL is a good technique for solving combinatorial optimization problems. The three algorithms reviewed in Section 3 all achieve good performance among the algorithms developed in those periods. Taking into account the selection bias when comparing algorithms, we can at least argue that the RL approach in combinatorial optimization achieves above-average performance. 

In addition to its performance, the RL approach has its own strengths. In the modern RL algorithms introduced in the recent five years, no human knowledge is required by those models. This means that the RL model starts from completely arbitrary values/states. Since deep RL has been widely used in approximating combinatorial optimization, the quality and capability of the RL approach have increased.

The potential of the RL approach approximation is greater than arithmetic or algorithmic approximation. With the rapid expansion of computing power, training times and the quantity of training data are considered less. The performance and capability of the RL approach approximation are thus continuously enhanced. This trend could become significant as quantum computing becomes more widely used in machine learning. 

Furthermore, if quantum computers are widely used in the future and fully developed, the RL approach could eventually fade out. Computers might then be able to directly compute the shortest paths of TSPs. Approximation algorithms will no longer be necessary as computing power is fully enhanced. However, the enhancement of computing power, as in quantum computing, requires a long time. We still need to focus on efficient ways of approximating the TSP. 

\bibliographystyle{IEEEtran}
\bibliography{IEEEfull}

\end{document}